\documentclass[lettersize,journal]{IEEEtran}
\usepackage{amsmath,amsfonts}
\usepackage{algorithmic}
\usepackage{array}
\usepackage[caption=false,font=normalsize,labelfont=sf,textfont=sf]{subfig}
\usepackage{textcomp}
\usepackage{stfloats}
\usepackage{url}
\usepackage{verbatim}
\usepackage{graphicx}
\usepackage{float}
\usepackage{placeins}    
\def\BibTeX{{\rm B\kern-.05em{\sc i\kern-.025em b}\kern-.08em
    T\kern-.1667em\lower.7ex\hbox{E}\kern-.125emX}}
\usepackage{balance}

\begin{document}

\title{A Backpropagation-Free Feedback-Hebbian Network \\ for Continual Learning Dynamics}

\author{
  \IEEEauthorblockN{Josh Li, Fow-Sen Choa}\\
  \IEEEauthorblockA{University of Maryland Baltimore County, Baltimore, MD\\
  Email: joshl1@umbc.edu, choa@umbc.edu}
}

\maketitle

\begin{abstract}
Feedback-rich neural architectures can regenerate earlier representations and inject temporal context, making them a natural setting for strictly local synaptic plasticity. Existing literature raises doubt about whether a minimal, backpropagation-free feedback--Hebbian system can already express interpretable continual-learning--relevant behaviors under controlled training schedules. In this work, we introduce a compact prediction-reconstruction architecture with a dedicated feedback pathway providing lightweight, locally trainable temporal context for continual adaptation. All synapses are updated by a unified local rule combining centered Hebbian covariance, Oja-style stabilization, and a local supervised drive where targets are available. With a simple two-pair association task, learning is characterized through layer-wise activity snapshots, connectivity trajectories (row/column means of learned weights), and a normalized retention index across phases. Under sequential $A\rightarrow B$ training, forward output connectivity exhibits a long-term depression (LTD)-like suppression of the earlier association while feedback connectivity preserves an $A$-related trace during acquisition of $B$. Under alternating sequence, both associations are concurrently maintained rather than sequentially suppressed. Architectural controls and rule-term ablations isolate the role of dedicated feedback in regeneration and co-maintenance alongside the role of the local supervised term in output selectivity and unlearning. Together, the results show that a compact feedback pathway trained with local plasticity can support regeneration and continual-learning--relevant dynamics in a minimal, mechanistically transparent setting.
\end{abstract}


\section{Introduction}

Feedback-rich neural architectures are attractive because they can regenerate earlier representations, inject top--down context, and propagate temporal information that stabilizes and accelerates adaptation \cite{2016_cortical_function}. Local update rules are suitable to this regime: plasticity occurs where information flows, avoiding explicit global credit signals, with Hebbian mechanisms providing a simple and well-studied substrate for feedback-driven modulation and input regeneration. In cortex-like recurrent hierarchies, the interplay of feedforward drive, feedback, and priors supports generative prediction, while population analyses suggest that task-relevant information can occupy a stable mnemonic subspace despite rich dynamics, where multiple circuit motifs (attractors, short-term synaptic processes, trained recurrent dynamics) can realize these functions \cite{2014_working_memory,2016_working_memory}.

A substantial body of work explores alternatives to conventional weight transport and global error propagation. Feedback Alignment (FA) showed that deep networks can learn useful mappings even when backward pathways use fixed random weights \cite{2016_feedback_bp}. Equilibrium Propagation (EqProp) casts learning in energy-based systems that use the same circuit in two phases, where a small output ``nudge'' yields a learning signal related to contrastive Hebbian learning and Spike-timing-dependent plasticity (STDP) \cite{2017_energy_based}. Predictive coding introduces error-encoding units and can approximate supervised learning with local plasticity under certain conditions \cite{2017_local_hb}. These directions primarily target credit assignment and biological plausibility; they are not typically framed around eliciting simple continual-learning (CL) primitives in a minimal feedback--Hebbian system. In contrast, mainstream CL methods are largely backpropagation-based and are commonly grouped into regularization, replay, and dynamic-architecture families \cite{2022_continual_learning}. Motivated by this gap, we raise a doubt about whether a compact feedback--Hebbian model trained with strictly local plasticity can express interpretable CL-relevant behaviors under controlled protocols.

This study proposes a backpropagation-free feedback--Hebbian network with dedicated feedforward and feedback pathways trained by a strictly local plasticity rule that combines centered Hebbian covariance, Oja-style stabilization, and a local supervised drive where targets are available. Our objective is mechanistic: we study a minimal two-layer feedforward pathway paired with a two-layer feedback pathway on a small, controllable two-pair association task. Rather than emphasizing benchmark accuracy, we characterize learning by tracking layer-wise activity and weight evolution across training phases using simple connectivity summaries.

Within this setting, we operationalize three continual-learning--relevant behaviors. (i) \emph{Retention under sequential training}: after learning association~A and then training with association~B only, A-related structure persists in the feedback pathway (as reflected by feedback-layer connectivity retention) even as the forward pathway adapts to express B at the output. (ii) \emph{LTD-like unlearning at the forward output stage}: sequential A$\rightarrow$B training suppresses A-specific forward output connectivity while strengthening B-specific connectivity, yielding a targeted weakening reminiscent of synaptic depression. (iii) \emph{Co-maintenance under temporal alternation}: under alternating sequence (e.g., $A,B,A,B,\ldots$), both associations are concurrently maintained, with task-relevant connectivity remaining elevated for both target sets rather than exhibiting sustained sequential suppression.

Our contributions include: (1) We introduce a minimal prediction--reconstruction architecture in which a dedicated feedback pathway is trained to reconstruct earlier-layer activity and is injected as additive temporal context. (2) We provide a unified local learning rule that is applied consistently to forward prediction and feedback reconstruction without weight transport or global error propagation. (3) Through layer-wise connectivity trajectories and phase-wise retention analyses, we show that this compact system expresses the defined continual-learning--relevant behaviors under sequential and interleaved training protocols.
\section{Methods}

\begin{figure*}[t]
    \centering
    \includegraphics[width=0.8\textwidth]{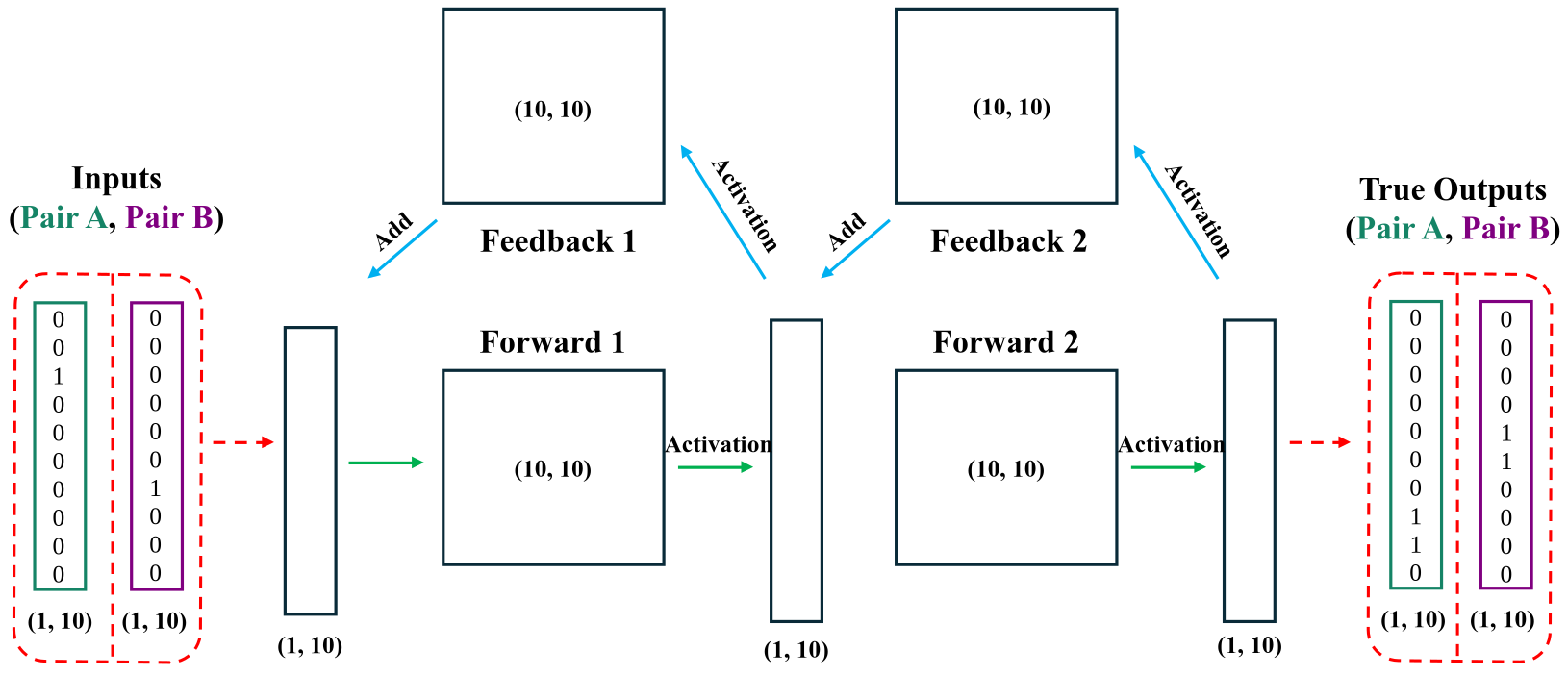}
    \caption{Backpropagation-free feedback-Hebbian network. Green: feedforward pathways; blue: dedicated feedback pathways. Feedforward and feedback use \emph{distinct} weight matrices; backward signals do not reuse feedforward weights in reverse.}
    \label{fig:model_structure}
\end{figure*}

\subsection{Task and Data Representation}
Our dataset is inspired by cortical stimulation paradigms but used here to create a small, controllable association task. Inputs and targets are $1{\times}10$ arrays corresponding to ten stimulation sites. Entries take values in $[0,1]$; in our experiments we use $\{0,1\}$ for clarity (0 = dormant, 1 = active), but the model does not require exact binary values.

We study two input positions (3, 7) and four target positions (5, 6, 8, 9) with fixed pairings: Pair A has position~3 that links to targets~8 and~9, and Pair B has position~7 links to targets~5 and~6. Concretely, when position~3 is set to 1 in the input, positions~8 and~9 are set to 1 in the target; when position~7 is set to 1 in the input, positions~5 and~6 are set to 1 in the target.

This yields a small multi-label mapping (multi-site input $\rightarrow$ multi-site output) that is easy to probe and visualize. We use it to generate the phase schedules, where we define the continual-learning diagnostics and their metrics, and to report the resulting weight/activation trajectories.

\subsection{Model Structure}
We instantiate the ten-site association task with a compact network that uses two forward layers and two dedicated feedback layers, each of size $10{\times}10$. The forward pathway computes the supervised input$\rightarrow$output mapping in two stages, and the feedback pathway provides a separate channel that reconstructs earlier activity and injects it as context at the next time step.

As shown in Fig.~\ref{fig:model_structure}, green arrows denote forward computation and blue arrows denote feedback. Anatomical directions are fixed: short-term adaptation changes synaptic \emph{efficacy} (weights), not wiring. More importantly, forward connections are not reused “in reverse”, as backward signals travel on their own feedback connections. This keeps the roles distinct: forward layers carry prediction toward the outputs, while feedback layers reconstruct earlier activity to modulate subsequent processing via a lightweight, additive pathway.

The temporal use of feedback proceeds as follows. Let $t$ index discrete steps. At $t{=}0$, no prior forward activity is available and feedback is inactive. For $t{\ge}1$, the previous step’s activity in the forward layer drives its paired feedback layer, forming a reconstruction of the earlier layer’s state, which is then added to the forward layer’s current drive. The merge is additive to preserve dimensionality and locality—no extra parameters or concatenation are introduced—yielding a lightweight local recurrence that carries recent context into the current computation.

We now justify the key structural choices, including depth, width, and the use of dedicated feedback. We use two forward layers to separate the two roles studied here: the second forward layer supports supervised prediction to the outputs, and the associated feedback reconstructs the first forward layer’s activity to stabilize and guide subsequent inputs. The width of ten units per layer aligns with the ten stimulation sites, keeping parameter counts small and making weight and activity trajectories easy to inspect. Accordingly, only the synaptic efficacies of these fixed connections are updated.

\subsection{Algorithm}
\label{subsec:algorithm}
We consider a local plasticity rule that trains the learned connections in Fig.~\ref{fig:model_structure}. Let $x\in\mathbb{R}^{10}$ denote the input to a trained layer and $y\in\mathbb{R}^{10}$ its output. Each trained weight matrix $W\in\mathbb{R}^{10\times 10}$ maps $x$ to $y$ (rows index output units; columns index input units). The same rule is applied to the two forward matrices and to the two dedicated feedback matrices. For layers without supervision we set $t_i{=}y_i$ such that the supervised term is neutral.

The synaptic update per epoch $i$ is
\begin{equation}
\label{eq:update}
\Delta w_i
= \mathrm{lr}\,\Bigl[
\bigl(x_i-\langle x_i\rangle\bigr)\bigl(y_i-\langle y_i\rangle\bigr)
- \beta\,\bigl(y_i-\langle y_i\rangle\bigr)^2\,w_i
+ (t_i-y_i)\,x_i
\Bigr],
\end{equation}
where $\mathrm{lr}$ is the learning rate, $\beta$ is the Oja-style normalization strength, and $x_i,y_i,t_i$ are the input, output, and (when present) target activities at epoch $i$ respectively. Activities are centered by exponential running means
\[
\langle x_i\rangle=(1-\alpha)\langle x_{i-1}\rangle+\alpha x_i,\qquad
\langle y_i\rangle=(1-\alpha)\langle y_{i-1}\rangle+\alpha y_i,
\]
where $\alpha$ is a constant smoothing parameter that sets the effective averaging window to approximately $1/\alpha$. The first (covariance) term updates weights based on deviations from average activity, preventing units with persistently high activity from dominating irrespective of input statistics \cite{hebb1949,bienenstock1982}. The second (Oja) term implements activity-dependent normalization that counteracts runaway Hebbian growth \cite{oja1982}. The final supervised drive is local and biases updates toward the provided targets via $(t_i-y_i)x_i$; when no supervision is provided for a layer, setting $t_i{=}y_i$ removes this contribution while leaving the other terms unchanged.

The activation function is
\[
\phi(z)=\tfrac{1}{2}\bigl(1+\tanh(z)\bigr)\in[0,1],
\]
where $z$ denotes the layer output pre-activation. The function matches the dataset range, provides smooth bounded responses, and improves stability for local Hebbian updates in the presence of feedback. Unbounded activations such as ReLU can amplify Hebbian growth under feedback unless additional controls are introduced.

Unless noted otherwise, we use $\alpha{=}0.01$, $\mathrm{lr}{=}0.001$, and $\beta{=}1$. Weights are initialized with a standard variance-scaled uniform distribution using fan-in scaling \cite{he2015delving}. Running means $\langle x\rangle$ and $\langle y\rangle$ are initialized to zero and carried across protocol phases to reflect ongoing baseline drift.

Centering directs plasticity to deviations from the current baseline, preventing persistently active units from dominating irrespective of input statistics and mitigating interference across phases. The Oja term constrains row norms and induces competition, enabling activity-dependent depression when output activity is high without matching input drive; this both prevents runaway growth and supports equalization during unlearning phases. The supervised drive accelerates acquisition while maintaining locality and leaving normalization intact. The bounded activation caps signal magnitudes, stabilizing dynamics when feedback injects recent activity. Updates are local and per synapse with $O(|W|)$ cost per step; no gradients or replay buffers are stored.

\subsection{Training Procedure}
\label{subsec:protocols}
Training uses the aforementioned local, backpropagation-free rule (Eq.~\eqref{eq:update}) to optimize forward prediction and feedback reconstruction. Weight transport is unnecessary, while all plasticity is local.

Supervision is applied only where targets exist: forward layer~2 is driven toward the task output vector, and each feedback layer is driven to reconstruct the input to its paired forward layer. Forward layer~1 receives no explicit target; its weights adapt locally from the incoming activity and Oja normalization. Eq. ~\eqref{eq:update} is applied once to each learned matrix with additive updates $w \leftarrow w + \Delta w$ using activities centered by exponential running means. The running means $\langle x\rangle$ and $\langle y\rangle$ persist across phases.

To elicit distinct associative behaviors, we use two regimes with matched total sample counts.

\textbf{Regime I: Sequential A then B (LTD-like effect and retention).}
Training proceeds in two phases. Phase 1 presents only pair~A samples (input site~3 mapping to output sites~8 and~9), 50 samples per epoch for 10 epochs. Phase 2 then presents only pair~B samples (input site~7 mapping to output sites~5 and~6), 50 samples per epoch for 10 epochs. Pair~A is not presented during Phase 2. This protocol tests depression of the previously learned association in the forward pathway during Phase 2, an LTD-like effect under the local rule, and retention of A-related connectivity in the feedback pathway.

\textbf{Regime II: Interleaved A/B (conditioning).}
Each epoch contains 100 samples arranged as a deterministic alternation $A,B,A,B,\ldots$ with 50 instances of each pair per epoch, starting with $A$. Training runs for 10 epochs, matching the total number of A and B presentations in Regime~I. This protocol tests associative linkage under close temporal alternation, consistent with classical conditioning: repeated alternation of distinct input-output mappings induces measurable cross-association between units involved in A and B while the primary mappings remain intact.

\subsection{Connectivity and Retention Metrics}

We use weight-derived metrics to summarize how synaptic strengths evolve over training and to obtain scalar measures that complement the connectivity trajectory plots. All metrics in this subsection are computed post hoc from the learned weight matrices and do not influence the network dynamics or the plasticity rule.

Each trained synaptic matrix $W\in\mathbb{R}^{10\times 10}$ has rows indexing output units and columns indexing input units. For a given matrix, \emph{input connectivity} is defined as the column-wise mean, yielding a $1{\times}10$ profile in which each entry reflects how strongly the corresponding input site projects into that layer on average. Conversely, \emph{output connectivity} is defined as the row-wise mean, yielding a $1{\times}10$ profile in which each entry reflects how strongly the corresponding output unit integrates its inputs on average. These input and output connectivity profiles are computed separately for each trained forward and feedback matrix for all training protocols. The resulting trajectories are then visualized to illustrate how synaptic strengths redistribute across sites and layers over time.

In addition to these trajectories, we use a normalized retention index to summarize how strongly a given connectivity measure changes during a later training phase, relative to the magnitude of the change induced by an earlier phase. Let $C(e)$ denote a scalar connectivity summary at epoch $e$ for a chosen layer.

Given an initial baseline epoch, an earlier training phase, and a subsequent training phase, we define three values $C^{0}$, $C^{\mathrm{pre}}$, and $C^{\mathrm{post}}$, where $C^{0}$ is the value at baseline, $C^{\mathrm{pre}}$ is the value at the end of the earlier phase, and $C^{\mathrm{post}}$ is the value at the end of the later phase. We then define the normalized index as
\begin{equation}
\label{eq:retention_index}
    R \;=\; \frac{C^{\mathrm{post}} - C^{\mathrm{pre}}}{\bigl|C^{\mathrm{pre}} - C^{0}\bigr|}.
\end{equation}
The numerator captures the signed change in connectivity during the later phase, while the denominator normalizes this change by the magnitude of the earlier-phase change relative to baseline. The absolute value ensures a nonnegative normalization scale so that the sign of $R$ reflects the direction of change from $C^{\mathrm{pre}}$ to $C^{\mathrm{post}}$. Values near $R{\approx}0$ indicate minimal change during the later phase (high retention). Positive values ($R{>}0$) indicate that connectivity increased during the later phase, whereas negative values ($R{<}0$) indicate a decrease. The magnitude $|R|$ quantifies the size of this change in units of the earlier-phase change.

When retention values are reported, the corresponding baseline, pre-phase, and post-phase epochs, as well as the underlying choice of connectivity summary $C(e)$, are specified alongside each protocol’s results.
\section{Results}
\label{sec:results}

In this experiment, we evaluate whether the feedback-Hebbian network exhibits three continual-learning-relevant primitives on the two-pair association task. These primitives are inspired by circuit-level phenomena in neuroscience \cite{duvarci2014amygdala, roy2017silent, dan2004stdp}, but are instantiated here as operational behaviors in a small, controlled supervised setting.

First, \emph{learning-without-forgetting} denotes retaining a trace of an earlier association in synaptic connectivity even after the network has adapted its output behavior to a newly trained association. Concretely, after sequential training on pair~A followed by pair~B, the network should produce the pair~B target at the output, while weight-derived measures indicate that A-related connectivity remains elevated relative to baseline, particularly in the feedback pathway.

Second, \emph{unlearning} refers to an LTD-like weakening of synapses that previously supported an association that is no longer produced at the output. In our setting, unlearning is expressed when sequential A$\rightarrow$B training reduces A-specific forward connectivity that drives the output, thereby suppressing the A-related output pattern while A-related traces can persist in the feedback layers.

Third, \emph{conditioning} captures the effect of temporally interleaving distinct associations. Under an alternating A/B schedule, the network should co-maintain both input--output mappings, reflected by sustained elevation of connectivity at both A- and B-related target sites.

Our analyses proceed in two stages. We first verify baseline activity-level behavior by checking that output activation concentrates on target sites and that feedback regenerates the corresponding input pattern from the output. We then interpret sequential and interleaved learning using the connectivity profiles and normalized retention index, which summarize how task-relevant synaptic strengths evolve from baseline through earlier and later training phases. Here, we apply this framework to single-association learning and regeneration, sequential A$\rightarrow$B training, interleaved A/B conditioning, architectural controls, and rule-term ablations.

\begin{figure}[!t]
\centering
\includegraphics[width=2.5in]{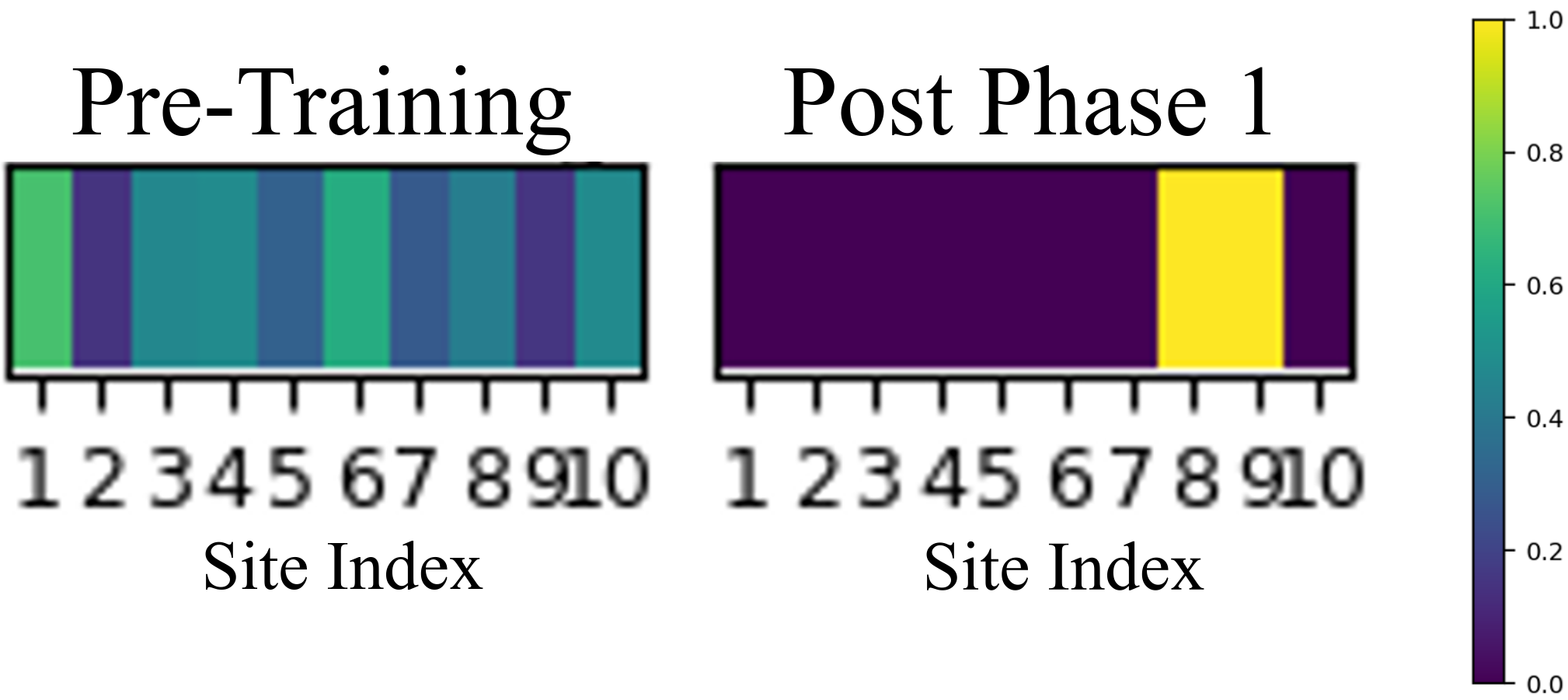}
\caption{Baseline forward prediction for pair A (input site 3 active). Output activity across the 10 sites (a) before training and (b) after Phase 1. Target output sites are 8 and 9.}
\label{fig:baseline_forward}
\end{figure}

\begin{figure}[!t]
\centering
\includegraphics[width=2.5in]{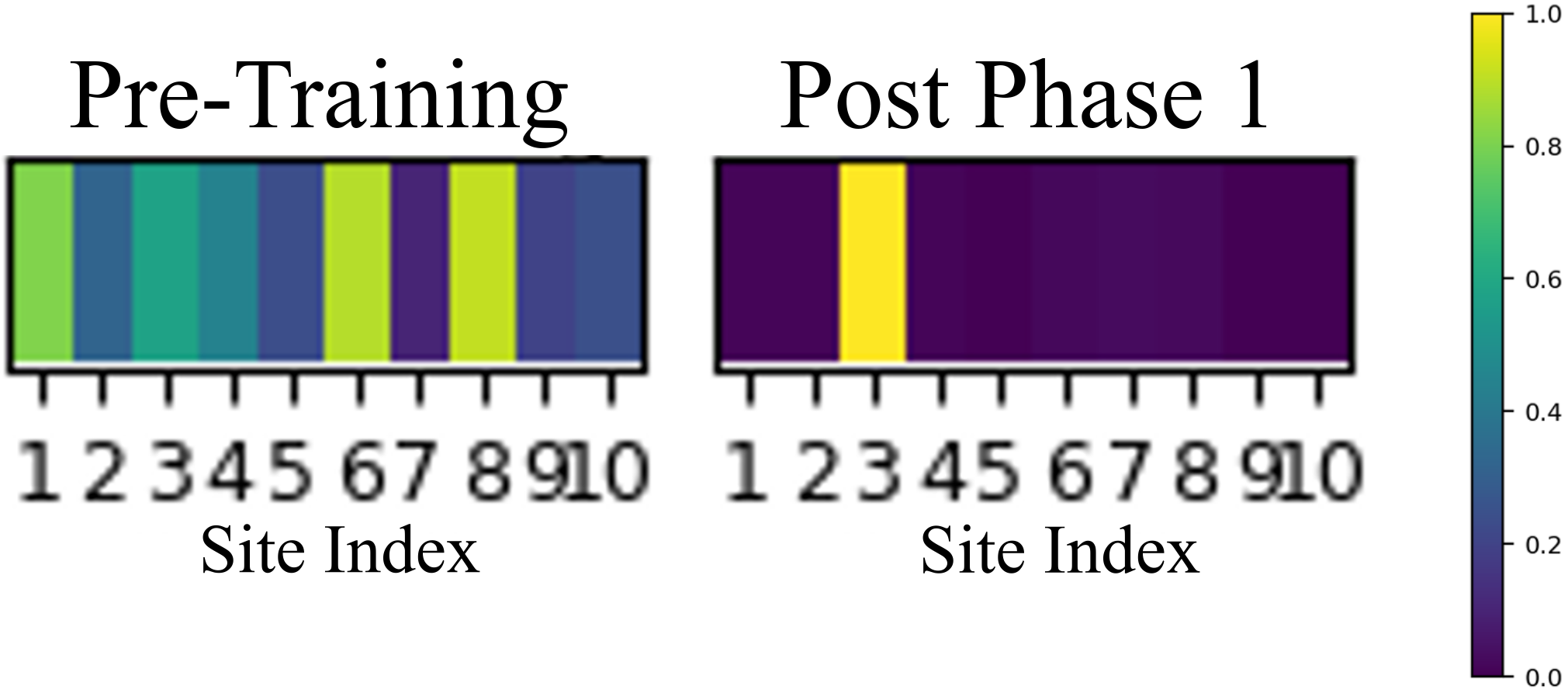}
\caption{Baseline feedback-driven regeneration for pair A. Regenerated input-side activity across the 10 sites when driven by the pair A output pattern (a) before training and (b) after Phase 1. The corresponding stimulated input site is 3.}
\label{fig:baseline_feedback}
\end{figure}

\subsection{Single-Association Learning and Regeneration}
\label{subsec:single_assoc}

We first verify the two baseline capabilities of the proposed 2FF+2FB network: (i) learning a single input--output association in the forward direction and (ii) regenerating the corresponding input pattern via the dedicated feedback pathway. The network is trained on pair~A only (Phase 1 of the sequential A-then-B training regime).

\paragraph{Forward prediction}
In the \emph{Pre-training} snapshot (random initialization), the output response to the pair~A input is unstructured, as shown in Fig.~\ref{fig:baseline_forward}. After completing Phase 1 training (\emph{Post Phase 1} snapshot), the same input elicits a selective output in which the dominant components coincide with the pair~A target sites (outputs~8 and~9), while non-target sites are driven toward zero.

\paragraph{Feedback-driven regeneration}
We then probe whether the feedback pathway can regenerate the input-side pattern associated with pair~A. When the feedback pathway is driven by the pair~A output pattern, the reconstructed activity is diffuse in the \emph{Pre-training} snapshot as shown in Fig.~\ref{fig:baseline_feedback}. After Phase 1 training (\emph{Post Phase 1} snapshot), the reconstruction is dominated by the originally stimulated input site~3, with non-target sites driven toward zero.

These baseline results verify single-association learning in the forward pathway and input regeneration through dedicated feedback, providing a reference point for the sequential and interleaved protocols analyzed next.

\begin{figure}[!t]
\centering
\includegraphics[width=\columnwidth]{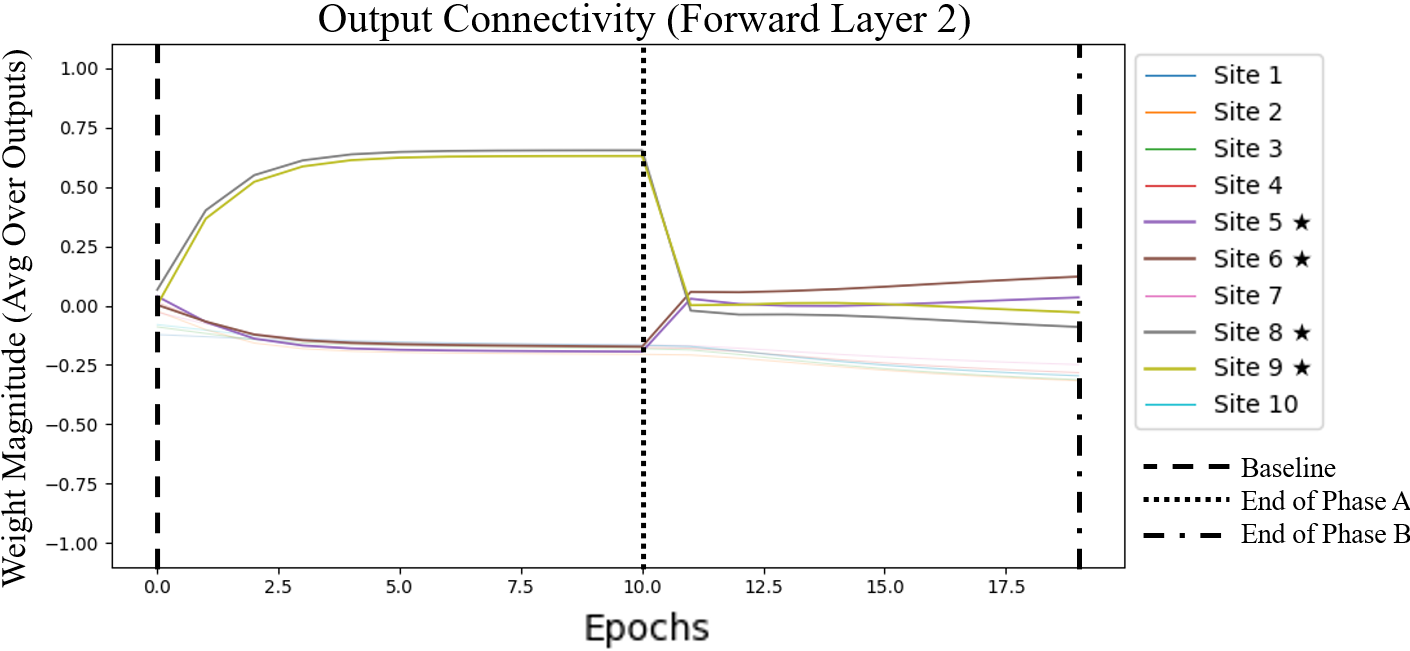}
\caption{Sequential A$\rightarrow$B training (Regime I): forward layer~2 output connectivity across epochs. Traces for the pair~B target outputs (sites~5-6) and the earlier pair~A targets (sites~8-9) are emphasized; other sites are shown faintly. The Phase 1/Phase 2 transition and the baseline/end-of-phase epochs ($e{=}0,10,20$) are marked.}
\label{fig:seq_forward_outconn}
\end{figure}

\begin{figure}[!t]
\centering
\includegraphics[width=\columnwidth]{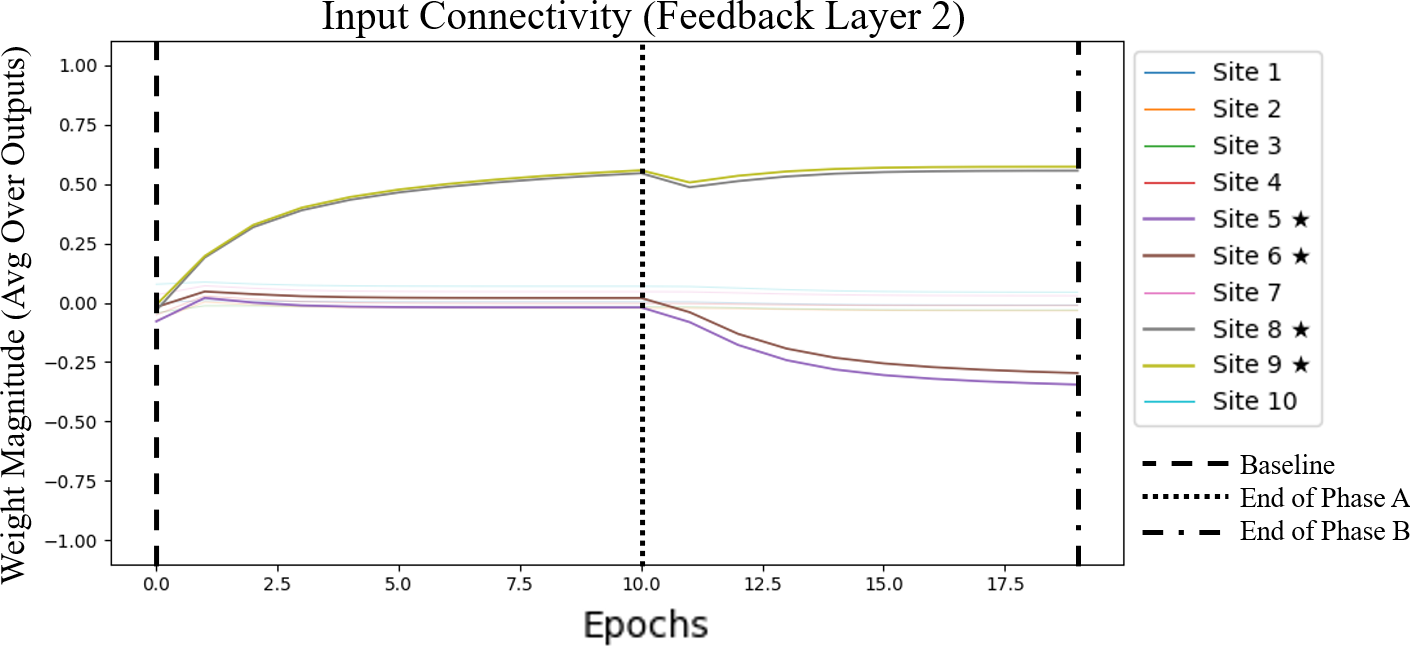}
\caption{Sequential A$\rightarrow$B training (Regime I): feedback layer~2 input connectivity across epochs. Traces corresponding to forward-output sites involved in pair~A (sites~8-9) are emphasized to assess retention during Phase 2; other sites are shown faintly. The Phase 1/Phase 2 transition and the baseline/end-of-phase epochs ($e{=}0,10,20$) are marked.}
\label{fig:seq_feedback_inconn}
\end{figure}

\subsection{Sequential A$\rightarrow$B: Retention and LTD-Like Unlearning}
\label{subsec:sequential_AB}

Next, we test unlearning and retention under the sequential A$\rightarrow$B protocol (Regime I). In the trajectory plots, the Phase 1/Phase 2 boundary and the baseline and endpoint epochs are marked on the highlighted traces.

\paragraph{LTD-like unlearning in the forward output stage}
Fig.~\ref{fig:seq_forward_outconn} plots the output connectivity of forward layer~2. During Phase 1, connectivity concentrates on the pair~A target outputs (sites~8 and~9). After switching to Phase 2, connectivity to the pair~B targets (sites~5 and~6) increases, while the earlier pair~A targets are driven down, consistent with LTD-like suppression at the output stage. This is summarized by the retention index (Eq.~\eqref{eq:retention_index}), where $R\approx 0$ indicates minimal Phase 2 change and the sign indicates increase/decrease: at the pair~B targets, $R_5{=}0.98$ and $R_6{=}1.68$, whereas at the pair~A targets, $R_8{=}{-}1.27$ and $R_9{=}{-}1.05$.

\paragraph{Retention of the earlier association in the feedback pathway}
Fig.~\ref{fig:seq_feedback_inconn} plots the input connectivity of feedback layer~2 over the same training sequence. In contrast to the forward output layer, the pair~A-related sites remain stable during Phase 2: $R_8{=}0.04$ and $R_9{=}0.05$, indicating minimal Phase 2 change relative to the Phase 1 change.

Sequential training therefore yields a separation of roles across pathways: the forward output stage shows an LTD-like reduction of the earlier association during Phase 2, while the feedback pathway preserves an A-related trace during acquisition of the new association.

\begin{figure}[!t]
\centering
\includegraphics[width=\columnwidth]{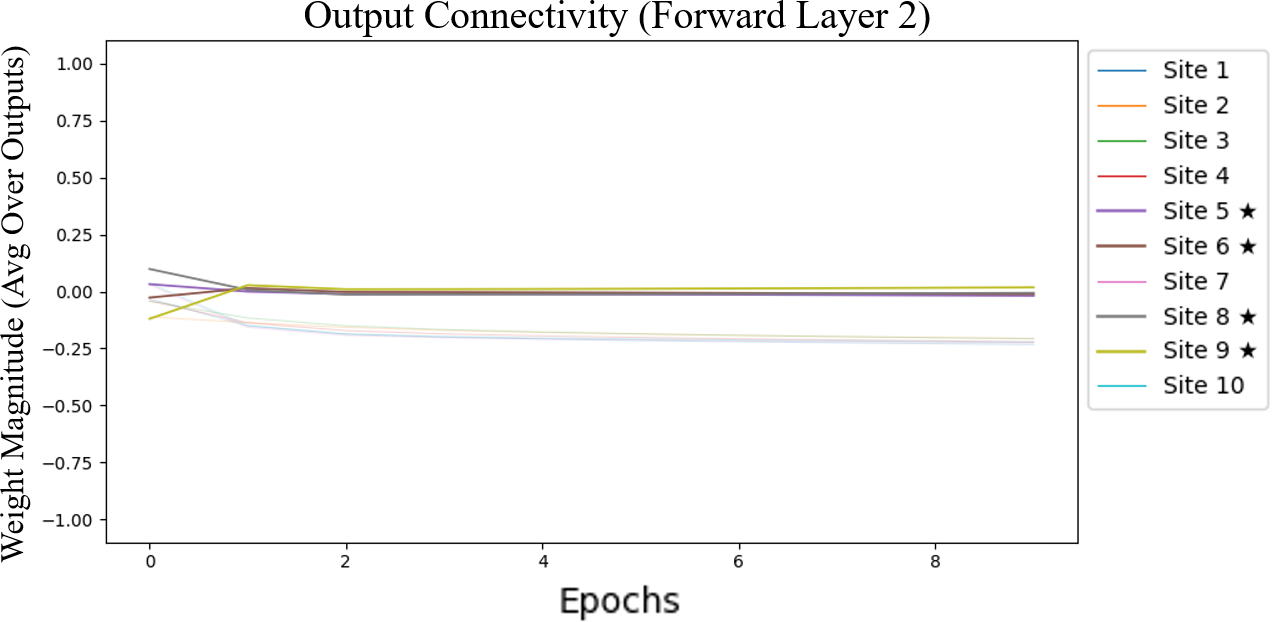}
\caption{Interleaved A/B conditioning: forward layer~2 output connectivity under Regime~II ($A,B,A,B,\ldots$ within each epoch). Connectivity to pair~A targets (sites~8 and~9) and pair~B targets (sites~5 and~6) increases and remains elevated relative to non-target sites, indicating concurrent maintenance of both associations.}
\label{fig:int_forward_outconn}
\end{figure}

\begin{figure}[!t]
\centering
\includegraphics[width=\columnwidth]{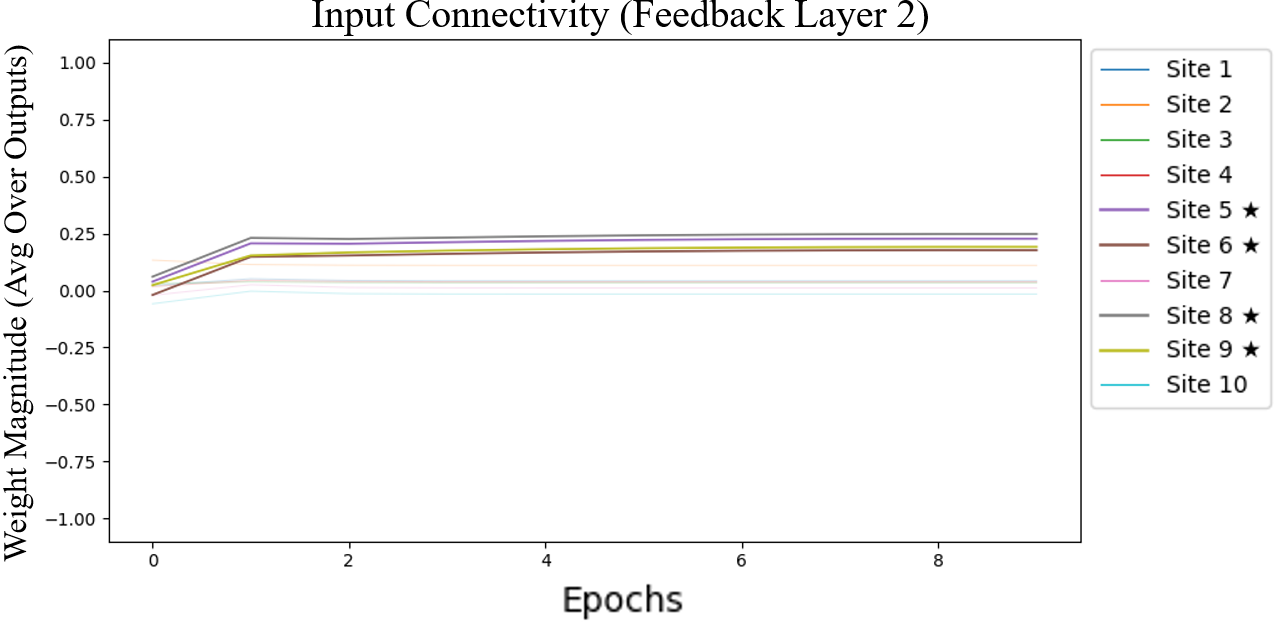}
\caption{Interleaved A/B conditioning: feedback layer~2 input connectivity under Regime~II ($A,B,A,B,\ldots$ within each epoch). Task-relevant sites show concurrent strengthening and stable persistence across training, consistent with co-encoding of both associations in the feedback pathway.}
\label{fig:int_feedback_inconn}
\end{figure}

\subsection{Interleaved A/B: Conditioning Under Temporal Alternation}
\label{subsec:interleaved_AB}

We then evaluate the conditioning primitive under temporal alternation (Regime~II), where samples from pair~A and pair~B are presented in a deterministic interleaving $A,B,A,B,\ldots$ within each epoch. In this setting, conditioning corresponds to concurrent acquisition and maintenance of both associations rather than sequential suppression of the earlier mapping.

\paragraph{Concurrent co-maintenance at the forward output stage}
Fig.~\ref{fig:int_forward_outconn} plots the output connectivity of forward layer~2 under interleaved training. Connectivity to the pair~A target outputs (sites~8 and~9) and to the pair~B target outputs (sites~5 and~6) increases during training and remains elevated relative to non-target sites. Unlike the sequential A$\rightarrow$B protocol analyzed above, the interleaved schedule does not induce a sustained depression of the earlier association; instead, both target sets are jointly maintained at the output stage.

\paragraph{Parallel co-learning in the feedback pathway}
Fig.~\ref{fig:int_feedback_inconn} shows the input connectivity of feedback layer~2 over the same interleaved sequence. The same task-relevant sites exhibit concurrent strengthening and stable persistence throughout training, indicating that the feedback pathway co-encodes both associations under alternation rather than preferentially reinforcing only the most recent mapping.

Together, these results show that temporal alternation supports concurrent encoding of both associations in the forward prediction and feedback pathways.

\begin{figure*}[t]
    \centering
    \includegraphics[width=0.8\textwidth]{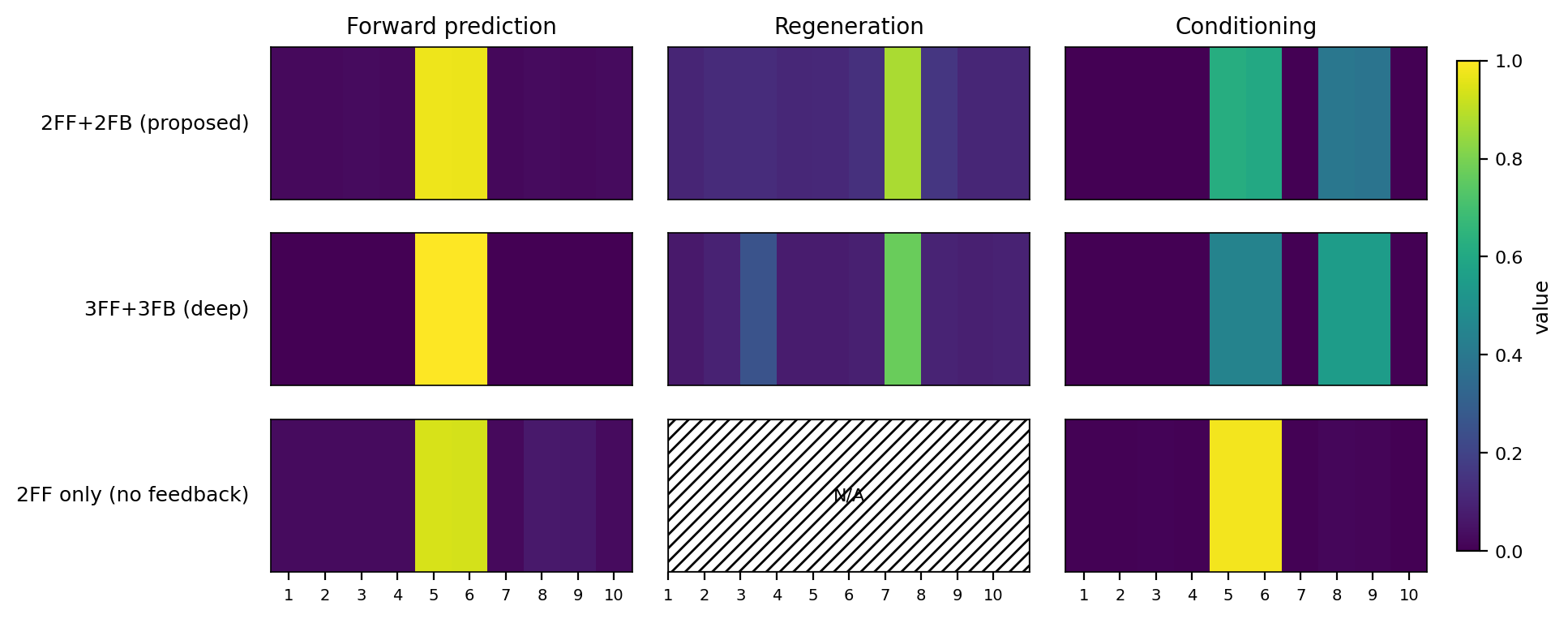}
    \caption{Architectural controls across depth and feedback pathways. Rows compare three architectures: 2FF+2FB (proposed), 3FF+3FB (deeper), and 2FF-only (forward pathway only, no feedback). Columns show three probes: \emph{Forward prediction} (output activity after sequential A$\rightarrow$B training, evaluated at the end of Phase 2), \emph{Regeneration} (input-side activity regenerated from the output via the feedback pathway after sequential A$\rightarrow$B training), and \emph{Conditioning} (output activity after interleaved A/B training, evaluated at the end of Regime~II). In the sequential probe, correct behavior emphasizes pair~B targets (outputs~5 and~6) with other sites near zero; in the conditioning probe, correct behavior maintains both pair~A targets (outputs~8 and~9) and pair~B targets (outputs~5 and~6) above non-target sites.}
    \label{fig:arch_grid}
\end{figure*}

\subsection{Architectural Controls: Depth and the Role of Feedback}
\label{subsec:architectural_controls}

Afterward, we evaluate architectural controls to assess whether the continual-learning--relevant primitives reported above depend on network depth and on the presence of a dedicated feedback pathway. Fig.~\ref{fig:arch_grid} summarizes three architectures across three probes.

The \emph{Forward prediction} and \emph{Regeneration} columns are evaluated after the sequential A$\rightarrow$B protocol at the end of Phase 2, so correct behavior is characterized by a selective expression of the pair~B mapping at the output and, when feedback is present, regeneration consistent with the pair~B association. The \emph{Conditioning} column is evaluated after the interleaved A/B protocol, where correct behavior requires concurrent maintenance of both pair~A and pair~B target outputs.

\paragraph{Proposed architecture (2FF+2FB) supports all probes}
The proposed 2FF+2FB architecture exhibits the expected behaviors across all columns as shown in Fig.~\ref{fig:arch_grid}. After sequential training, forward prediction at the end of Phase 2 emphasizes the pair~B target outputs (sites~5 and~6) while suppressing non-target sites, and the regeneration probe yields a selective reconstructed pattern consistent with the pair~B association. Under interleaved training, the conditioning probe shows concurrent expression of both pair~A targets (sites~8 and~9) and pair~B targets (sites~5 and~6), consistent with co-maintenance under temporal alternation.

\paragraph{Deeper architecture (3FF+3FB) is not necessary on this task}
The 3FF+3FB architecture produces qualitatively similar outcomes to 2FF+2FB across the same three probes, as also described in Fig.~\ref{fig:arch_grid}. For the present two-pair dataset, additional depth does not yield an observable change in the forward prediction, regeneration, or conditioning behaviors, supporting the use of 2FF+2FB as a sufficient configuration.

\paragraph{Removing feedback (2FF-only) breaks regeneration and conditioning}
Finally, the 2FF-only architecture isolates the role of feedback. As expected, the regeneration probe fails due to the absence of a feedback pathway. More importantly, the conditioning probe can no longer maintain both associations: after interleaved training, one target set dominates while the other collapses toward non-target levels as highlighted in Fig.~\ref{fig:arch_grid}. Notably, post Phase 2 forward prediction can still emphasize the pair~B targets, indicating that the forward pathway alone can express the most recently trained association, whereas dedicated feedback is required for regeneration and for stabilizing concurrent associations under temporal alternation.

In short, these architectural controls indicate that two forward layers with dedicated feedback are sufficient for the present task; removing feedback selectively abolishes regeneration and conditioning.

\subsection{Rule-Term Ablations: Covariance, Oja Normalization, and Supervision}
\label{subsec:rule_ablations}

Lastly, we ablate individual terms in the local update rule (Eq.~\eqref{eq:update}) to assess which components are necessary for the continual-learning--relevant primitives observed under the full rule. As references for the desired behaviors, sequential A$\rightarrow$B training under the proposed update exhibits LTD-like suppression of the earlier association at the forward output stage with retention in feedback as depicted in Figs.~\ref{fig:seq_forward_outconn},~\ref{fig:seq_feedback_inconn}, and interleaved training maintains both associations concurrently as highlighted in Figs.~\ref{fig:int_forward_outconn},~\ref{fig:int_feedback_inconn}.

\begin{table}[!t]
\centering
\caption{Peak synaptic magnitude at the end of sequential A$\rightarrow$B training (Regime I; end of Phase 2). For each update variant, we report the maximum absolute weight across all learned matrices and the layer in which this maximum occurs.}
\label{tab:max_weight_ablation}
\begin{tabular}{lcc}
\hline
\textbf{Update variant} & $\max |W|$ & \textbf{Layer of max} \\
\hline
Proposed (full rule) & 1.16 & Forward layer~1 \\
No decay (no Oja term) & 1.72 & Forward layer~1 \\
No covariance term & 1.99 & Forward layer~2 \\
No supervised term & 0.54 & Forward layer~1 \\
\hline
\end{tabular}
\end{table}

\paragraph{Weight-scale effects of normalization and centering}
Table~\ref{tab:max_weight_ablation} reports the maximum absolute weight across all learned matrices at the end of sequential A$\rightarrow$B training (end of Phase 2) for four update variants. Removing the Oja-style decay term increases the peak weight magnitude relative to the full rule (1.72 vs.\ 1.16), and removing the covariance (centering) term yields the largest peak magnitude (1.99). In our experiments, these larger weight scales coincide with slower suppression of earlier-association connectivity during Phase 2 (as reflected by the forward output connectivity trajectories), and are consistent with the common view that larger synaptic gains can make local learning dynamics harder to reverse and potentially less stable, particularly in feedback-driven settings.

\begin{figure}[!t]
\centering
\includegraphics[width=\columnwidth]{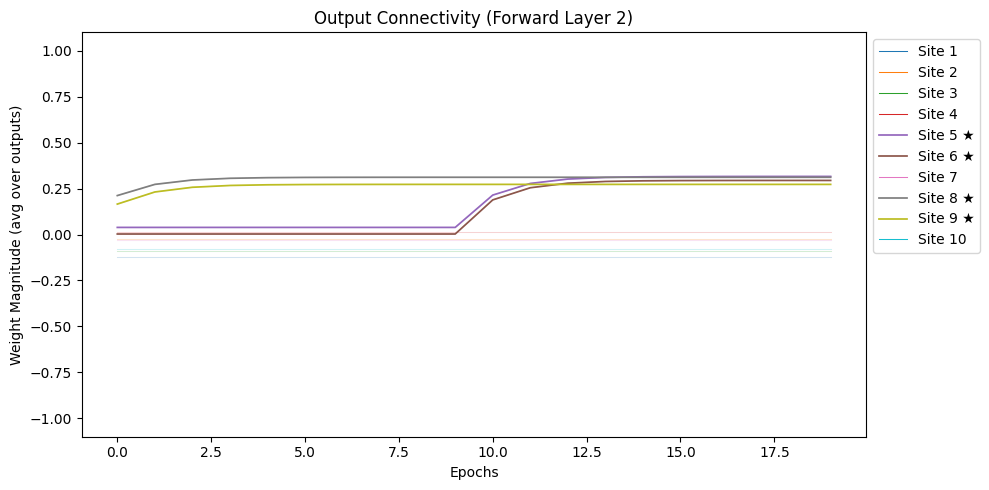}
\caption{Rule-term ablation (no supervised drive): sequential A$\rightarrow$B training (Regime I) forward layer~2 output connectivity across epochs. Unlike the full rule in Fig.~\ref{fig:seq_forward_outconn}, the earlier pair~A target outputs (sites~8-9) are not effectively suppressed during Phase 2 while pair~B targets (sites~5-6) are learned, indicating failure of LTD-like unlearning without the local supervised term. The Phase 1/Phase 2 transition and the baseline/end-of-phase epochs ($e{=}0,10,20$) are marked.}
\label{fig:abl_nosup_seq_forward_outconn}
\end{figure}

\begin{figure}[!t]
\centering
\includegraphics[width=\columnwidth]{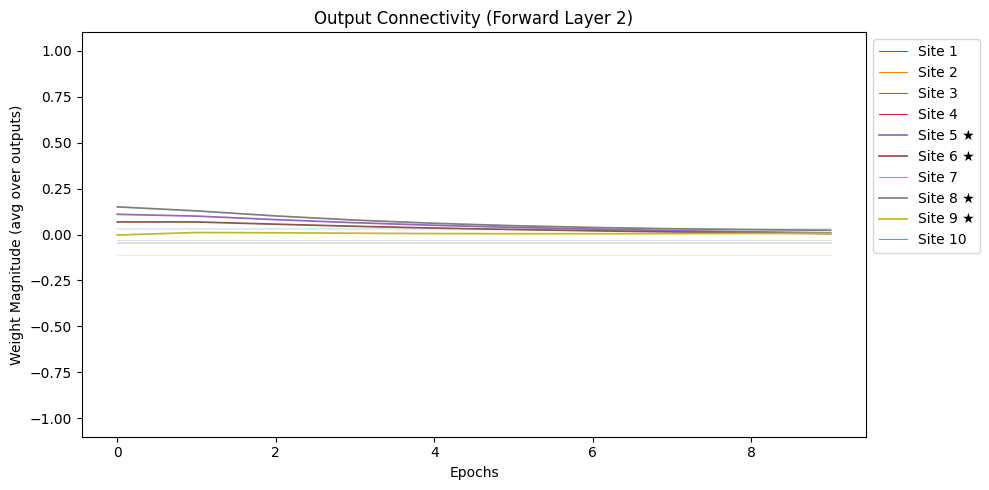}
\caption{Rule-term ablation (no supervised drive): interleaved A/B conditioning (Regime~II; $A,B,A,B,\ldots$ within each epoch) forward layer~2 output connectivity across epochs. In contrast to the full rule in Fig.~\ref{fig:int_forward_outconn}, the network fails to maintain both pair~A targets (sites~8-9) and pair~B targets (sites~5-6) concurrently under alternation, consistent with impaired suppression of non-target outputs when the supervised term is removed.}
\label{fig:abl_nosup_int_forward_outconn}
\end{figure}

\paragraph{Removing the supervised term disrupts unlearning under sequential training}
We then remove the local supervised drive $(t_i-y_i)x_i$ from Eq.~\eqref{eq:update} while keeping the covariance and Oja terms unchanged. Under sequential A$\rightarrow$B training, the forward layer~2 output connectivity no longer shows a clean LTD-like reduction of the earlier association during Phase 2 as plotted in Fig.~\ref{fig:abl_nosup_seq_forward_outconn}. In contrast to the full rule as depicted in Fig.~\ref{fig:seq_forward_outconn}, the earlier pair~A target outputs (sites~8--9) are not effectively driven down while pair~B targets (sites~5--6) are acquired. This indicates that Hebbian covariance updates with normalization are not sufficient on their own to suppress non-target outputs at the output stage; the supervised term provides the local error-driven component needed to selectively depress activity inconsistent with the current targets.

\paragraph{Removing the supervised term also impairs conditioning under interleaving}
The same ablation also disrupts conditioning under temporal alternation. Under the interleaved A/B protocol (Regime~II), forward layer~2 output connectivity fails to maintain both target sets concurrently when the supervised term is removed  as shown in Fig.~\ref{fig:abl_nosup_int_forward_outconn}. Whereas the full rule preserves elevated connectivity at both pair~A targets (sites~8--9) and pair~B targets (sites~5--6) throughout training as compared in Fig.~\ref{fig:int_forward_outconn}, removing supervision leads to dominance or interference that prevents stable co-maintenance of the two associations. Together with the sequential results, this supports the functional role of the supervised drive as the mechanism that enforces output selectivity---enabling LTD-like unlearning of outdated associations and supporting concurrent conditioning under alternation.

Overall, these ablations indicate complementary roles for the update terms: covariance and Oja-style normalization constrain synaptic scale and support stable adaptation, while the supervised term is required to selectively suppress non-target outputs and thereby realize both unlearning under sequential training and conditioning under temporal alternation.
\section{Conclusions}
We presented a compact, backpropagation-free feedback--Hebbian network that couples supervised forward prediction with a dedicated feedback pathway trained for reconstruction and used as lightweight temporal context. Both pathways are learned with a unified local plasticity rule that combines centered Hebbian covariance, Oja-style stabilization, and a local supervised drive where targets are available. Rather than emphasizing benchmark-scale accuracy, we used a small, controllable two-pair association task to make layer-wise learning dynamics directly observable through activity snapshots, connectivity trajectories, and phase-wise retention summaries.

Across two training regimes, the model exhibited three operational continual-learning--relevant primitives. First, under sequential $A\rightarrow B$ training, forward layer~2 output connectivity shifted from the pair~A to pair~B, while the earlier pair~A structure was selectively suppressed, consistent with an LTD-like unlearning effect at the output stage. Secondly, in the same sequential protocol, feedback layer~2 preserved an A-related trace: input connectivity at the pair~A-related sites remained near-stable during Phase~B, indicating retention despite the forward pathway’s adaptation to the newly trained mapping. Lastly, under alternating sequence, both associations were concurrently maintained at the forward output stage, with parallel strengthening and persistence in the feedback pathway, consistent with conditioning under temporal alternation rather than sequential suppression.

Architectural controls and rule-term ablations further sharpened the mechanistic interpretation. The proposed 2FF+2FB configuration reliably expressed forward prediction, feedback-driven regeneration, sequential unlearning with retention, and interleaved co-maintenance. Removing the feedback pathway abolished regeneration and degraded stable co-maintenance under alternation, isolating dedicated feedback as the component that enables regeneration and supports concurrent associations. At the learning-rule level, covariance and Oja-style stabilization regulated synaptic scale and supported stable adaptation, while removing the local supervised drive disrupted output selectivity, impairing both LTD-like unlearning under sequential training and co-maintenance under alternating sequence. In short, these findings support a division of labor in which forward layers express the currently trained output mapping, while feedback layers can preserve and propagate association-related traces that are not forced to mirror the immediate output targets.

\section{Outlook}
This study is intentionally minimal, and several limitations motivate future work. The present task comprises two low-dimensional associations and short training horizons; it therefore remains unclear how the same local learning dynamics scale to higher-dimensional inputs, longer task sequences, structured task overlap, and noisier settings. A natural next step is to extend the same controlled sequential and interleaved protocols to richer multi-class or multi-label datasets while retaining diagnostics that localize where retention and interference emerge across layers and pathways. Beyond scaling, dedicated feedback trained by strictly local rules provides a controlled design space for testing additional biologically motivated motifs (e.g., alternative reconstruction targets, partial or gated feedback, and sparsity constraints) without introducing global gradient transport. 

In the present architecture, each feedback pathway is trained to reconstruct the activity of its paired forward layer, including reconstruction of the input representation as a special case. This explicit reconstruction signal suggests a practical role in settings that require online adaptation under locality constraints: feedback can carry short-horizon context and provide internal consistency signals while synaptic updates remain strictly local and do not require backpropagation through time. A longer-term extension is to couple this reconstruction pathway to a persistent storage mechanism, so that previously encountered internal states can be reinstated from partial cues across extended time—an episodic-memory-like operation. Importantly, the current model does not implement storage, temporal indexing, or capacity control; evaluating this direction will require sequence-level protocols (e.g., delayed recall and replay-like consolidation) and longer-horizon measurements of stability-plasticity trade-offs under sustained adaptation.

Another direction is hybridization with gradient-based training. A conservative use is to treat local feedback-reconstruction training as an initialization or auxiliary pretraining phase that shapes representations and stabilizes dynamics, followed by conventional gradient-based optimization for task-level performance. More broadly, the reconstruction pathway could also be leveraged as an auxiliary objective or regularizer for a separate learner, but the appropriate scheduling and the regimes in which hybrid training is beneficial remain open questions.

\bibliographystyle{IEEEtran}
\bibliography{bibliography}

@article{2014_working_memory,
title = {Working models of working memory},
journal = {Current Opinion in Neurobiology},
volume = {25},
pages = {20-24},
year = {2014},
note = {Theoretical and computational neuroscience},
issn = {0959-4388},
doi = {https://doi.org/10.1016/j.conb.2013.10.008},
url = {https://www.sciencedirect.com/science/article/pii/S0959438813002158},
author = {Omri Barak and Misha Tsodyks}
}

@article{
2016_working_memory,
author = {John D. Murray  and Alberto Bernacchia  and Nicholas A. Roy  and Christos Constantinidis  and Ranulfo Romo  and Xiao-Jing Wang },
title = {Stable population coding for working memory coexists with heterogeneous neural dynamics in prefrontal cortex},
journal = {Proceedings of the National Academy of Sciences},
volume = {114},
number = {2},
pages = {394-399},
year = {2017},
doi = {10.1073/pnas.1619449114},
URL = {https://www.pnas.org/doi/abs/10.1073/pnas.1619449114},
eprint = {https://www.pnas.org/doi/pdf/10.1073/pnas.1619449114}}

@article{
2016_cortical_function,
author = {David J. Heeger },
title = {Theory of cortical function},
journal = {Proceedings of the National Academy of Sciences},
volume = {114},
number = {8},
pages = {1773-1782},
year = {2017},
doi = {10.1073/pnas.1619788114},
URL = {https://www.pnas.org/doi/abs/10.1073/pnas.1619788114},
eprint = {https://www.pnas.org/doi/pdf/10.1073/pnas.1619788114}}

@article{2017_local_hb,
title={An approximation of the error backpropagation algorithm in a predictive coding network with local Hebbian synaptic plasticity},
author={Whittington, J. C. R. and Bogacz, R.},
journal={Neural Computation},
volume={29},
number={5},
pages={1229--1262},
year={2017},
doi={10.1162/NECO_a_00949}
}

@misc{2017_energy_based,
title={Equilibrium Propagation: Bridging the Gap Between Energy-Based Models and Backpropagation}, 
author={Benjamin Scellier and Yoshua Bengio},
year={2017},
eprint={1602.05179},
archivePrefix={arXiv},
primaryClass={cs.LG},
url={https://arxiv.org/abs/1602.05179}, 
}

@article{2016_feedback_bp,
title={Random synaptic feedback weights support error backpropagation for deep learning},
author={Lillicrap, Timothy P. and Cownden, Daniel and Tweed, Douglas B. and Akerman, Colin J.},
journal={Nature Communications},
volume={7},
pages={13276},
year={2016},
doi={10.1038/ncomms13276}
}

@article{2022_continual_learning,
title = {A wholistic view of continual learning with deep neural networks: Forgotten lessons and the bridge to active and open world learning},
journal = {Neural Networks},
volume = {160},
pages = {306-336},
year = {2023},
issn = {0893-6080},
doi = {https://doi.org/10.1016/j.neunet.2023.01.014},
url = {https://www.sciencedirect.com/science/article/pii/S089360802300014X},
author = {Martin Mundt and Yongwon Hong and Iuliia Pliushch and Visvanathan Ramesh}
}

@book{hebb1949,
  title        = {The Organization of Behavior: A Neuropsychological Theory},
  author       = {Hebb, Donald O.},
  year         = {1949},
  publisher    = {Wiley},
  address      = {New York}
}

@article{bienenstock1982,
  title        = {Theory for the development of neuron selectivity: orientation specificity and binocular interaction in visual cortex},
  author       = {Bienenstock, Elie L. and Cooper, Leon N. and Munro, Paul W.},
  journal      = {The Journal of Neuroscience},
  volume       = {2},
  number       = {1},
  pages        = {32--48},
  year         = {1982},
  doi          = {10.1523/JNEUROSCI.02-01-00032.1982}
}

@article{oja1982,
  title        = {Simplified neuron model as a principal component analyzer},
  author       = {Oja, Erkki},
  journal      = {Journal of Mathematical Biology},
  volume       = {15},
  pages        = {267--273},
  year         = {1982},
  doi          = {10.1007/BF00275687}
}

@inproceedings{he2015delving,
  title={Delving Deep into Rectifiers: Surpassing Human-Level Performance on ImageNet Classification},
  author={He, Kaiming and Zhang, Xiangyu and Ren, Shaoqing and Sun, Jian},
  booktitle={Proceedings of the IEEE International Conference on Computer Vision (ICCV)},
  year={2015}
}

@article{duvarci2014amygdala,
  title        = {Amygdala Microcircuits Controlling Learned Fear},
  author       = {Duvarci, Sevil and Par{\'e}, Denis},
  journal      = {Neuron},
  volume       = {82},
  number       = {5},
  pages        = {966--980},
  year         = {2014},
  doi          = {10.1016/j.neuron.2014.04.042}
}

@article{roy2017silent,
  title        = {Silent memory engrams as the basis for retrograde amnesia},
  author       = {Roy, Dheeraj S. and Muralidhar, Shruti and Smith, Lillian M. and Tonegawa, Susumu},
  journal      = {Proceedings of the National Academy of Sciences of the United States of America},
  volume       = {114},
  number       = {46},
  pages        = {E9972--E9979},
  year         = {2017},
  doi          = {10.1073/pnas.1714248114}
}

@article{dan2004stdp,
  title        = {Spike Timing-Dependent Plasticity of Neural Circuits},
  author       = {Dan, Yang and Poo, Mu-ming},
  journal      = {Neuron},
  volume       = {44},
  number       = {1},
  pages        = {23--30},
  year         = {2004},
  doi          = {10.1016/j.neuron.2004.09.007}
}

\end{document}